\documentclass[runningheads]{llncs}
\usepackage[T1]{fontenc}
\usepackage{graphicx,verbatim}
\usepackage{multirow}
\usepackage{booktabs}
\usepackage{arydshln}
\usepackage{amsmath}      
\usepackage{amssymb}       
\usepackage{amsfonts}   
\usepackage{bm}        
\usepackage[linesnumbered,ruled,vlined]{algorithm2e} 
\usepackage[table]{xcolor}
\usepackage{colortbl}
\usepackage[table]{xcolor}
\usepackage{colortbl}
\usepackage[table]{xcolor}
\usepackage{xurl} 
\usepackage[misc]{ifsym}     
\usepackage[colorlinks, urlcolor=blue]{hyperref} 
\SetKwInOut{KwInput}{Input}
\SetKwInOut{KwOutput}{Output}
\SetKwInOut{KwParameter}{Parameters}
\LinesNotNumbered

\definecolor{dgreen}{RGB}{0,150,0}
\definecolor{gray}{RGB}{127,127,127}
\definecolor{orange}{RGB}{255,120,0}

\newif\ifdraft
\drafttrue

\ifdraft
  \newcommand{\JC}[1]{{\color{red}{\bf JC: #1}}}
  
  \newcommand{\NB}[1]{{\color{dgreen}{\bf NB: #1}}}

\else
  \newcommand{\JC}[1]{}
  
  \newcommand{\NB}[1]{}
  
\fi

\newcommand{\parag}[1]{\vspace{-3mm}\paragraph{#1}}

\newcommand{\method}{$PlaneCycle$}

\newcommand{\algcomment}[1]{\textcolor{black!50}{\footnotesize #1}}

\newcommand{\bXi}{\bm{X}_{i}}
\newcommand{\bGi}{\bm{G}_{i}}
\newcommand{\cFt}{\mathcal{F}_{\theta}}
\newcommand{\bXo}{\bm{X}_{o}}
\newcommand{\bGo}{\bm{G}_{o}}
\newcommand{\bbR}{\mathbb{R}}

\begin{document}
\title{\method{}: Training-Free 2D-to-3D Lifting of Foundation Models Without Adapters}

\titlerunning{\method{}: Training-Free 2D-to-3D Model Lifting}

\author{
Yinghong Yu\inst{1,2}\textsuperscript{$\star$\orcidID{0009-0001-6382-0001}} \and
Guangyuan Li\inst{1,2}\textsuperscript{$\star$\orcidID{0009-0005-8063-7873}} \and
Jiancheng Yang\inst{1,2}\textsuperscript{$\star\star$\orcidID{0000-0003-4455-7145}}
}

\authorrunning{Y. Yu et al.}

\institute{
ELLIS Institute Finland, Espoo, 02150, Finland
\and
Aalto University, Espoo, 02150, Finland
\\
\email{jiancheng.yang@aalto.fi}
}

\maketitle

{\let\thefootnote\relax\footnotetext{$^\star \phantom{\star}$ Equal contribution. }}

{\let\thefootnote\relax\footnotetext{$^{\star\star}$ Corresponding author: Jiancheng Yang (\href{mailto:jiancheng.yang@aalto.fi}{jiancheng.yang@aalto.fi}).}}

\begin{abstract}


Large-scale 2D foundation models exhibit strong transferable representations, yet extending them to 3D volumetric data typically requires retraining, adapters, or architectural redesign. We introduce \method{}, a training-free, adapter-free operator for architecture-agnostic 2D-to-3D lifting of foundation models. \method{} reuses the original pretrained 2D backbone by cyclically distributing spatial aggregation across orthogonal $HW$, $DW$, and $DH$ planes throughout network depth, enabling progressive 3D fusion while preserving pretrained inductive biases. The method introduces no additional parameters and is applicable to arbitrary 2D networks. Using pretrained DINOv3 models, we evaluate \method{} on six 3D classification and three 3D segmentation benchmarks. Without any training, the lifted models exhibit intrinsic 3D fusion capability and, under linear probing, outperform slice-wise 2D baselines and strong 3D counterparts, approaching the performance of fully trained models. With full fine-tuning, \method{} matches standard 3D architectures, highlighting its potential as a seamless and practical 2D-to-3D lifting operator. These results demonstrate that 3D capability can be unlocked from pretrained 2D foundation models without structural modification or retraining. 
Code is available at \url{https://github.com/HINTLab/PlaneCycle}.



\end{abstract}

\section{Introduction}

Large-scale 2D foundation models~\cite{caron2021emerging,he2022masked,kirillov2023segment,simeoni2025dinov3} have demonstrated strong robustness and transferable representations across diverse domains, and have shown increasing promise in medical imaging~\cite{ma2024segment,liu2025does}, where data are limited and heterogeneous~\cite{yang2024multi}. While transferring pretrained 2D models to downstream 2D tasks is straightforward via fine-tuning or low-rank adaptation~\cite{hu2022lora}, extending them to volumetric 3D data is considerably less natural, despite many clinical imaging modalities (e.g., CT, MRI, OCT) being inherently 3D.

A common strategy applies 2D models slice-by-slice and aggregates predictions post hoc~\cite{liu2025does}, which is computationally efficient but neglects cross-slice dependencies. Alternatively, 2D backbones can be converted into full 3D~\cite{carreira2017quo,arnab2021vivit} or augmented with adapters~\cite{yang2021asymmetric,wu2025medical,liu2025revisiting}. Notably, such converted models typically exhibit no intrinsic 3D capability prior to 3D retraining. Other efforts rely on large-scale video pretraining~\cite{wang2024internvideo2} or dedicated 3D medical pretraining~\cite{hamamci2026generalist,wu2025large}, yet the diversity and accessibility of these data sources remain significantly more limited than natural 2D imagery. In addition, given the enormous computational investment behind modern 2D foundation models, DINOv3 reports 9M H100 GPU hours and 2600 tCO2eq~\cite{simeoni2025dinov3}, developing mechanisms that more effectively reuse pretrained 2D representations is essential for sustainability.

These observations raise a fundamental question: can 3D capability emerge from pretrained 2D foundation models \textit{without} modifying their architecture or parameters? Ideally, such a mechanism should be training-free while remaining fully compatible with fine-tuning when supervision is available. The closest related perspective is ACS convolution~\cite{yang2021reinventing}, which reorganizes conv kernels across orthogonal planes; however, it is restricted to CNN architectures. In an era where ViT~\cite{dosovitskiyimage} foundation models increasingly dominate, there is a need for architecture-agnostic operators capable of reusing both modern ViT and established CNN backbones~\cite{isensee2024nnu,roy2025mednext} within a unified 2D-to-3D lifting framework.

\begin{figure}[tb]
\centering
\includegraphics[width=1\textwidth]{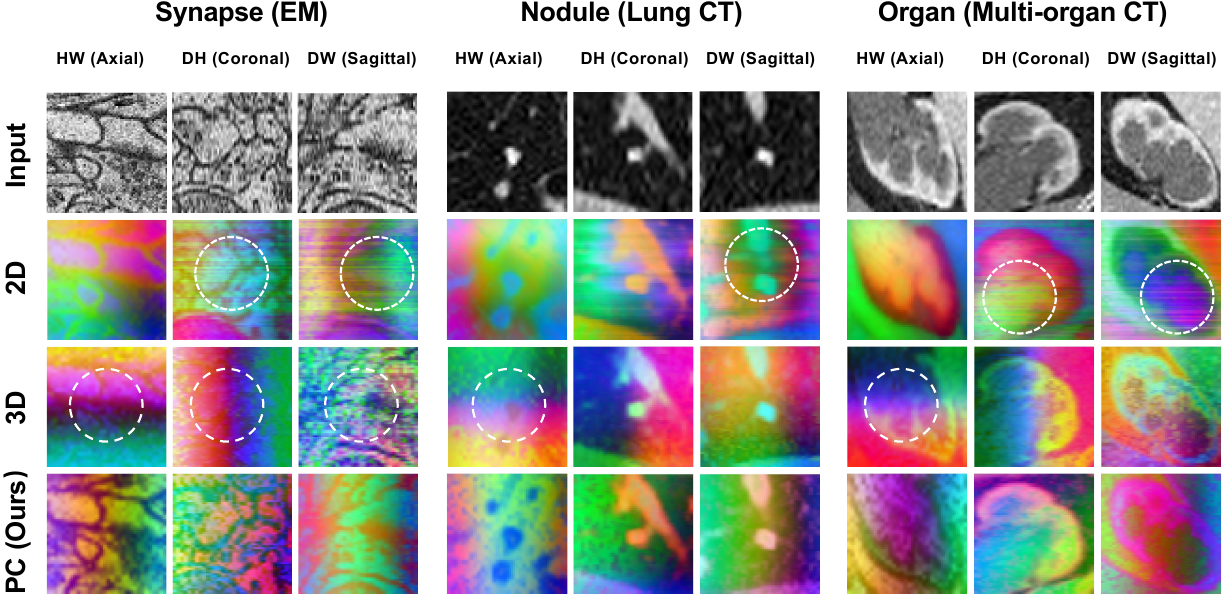}
\caption{
PCA visualizations of frozen lifted DINOv3~\cite{simeoni2025dinov3} features on three 3D datasets~\cite{medmnistv2} across $HW$, $DW$, and $DH$ planes; inconsistencies circled.
} \label{fig:feature_visualization}
\end{figure}

To address this question, we propose \method{}, a parameter-free operator for architecture-agnostic 2D-to-3D lifting. \method{} preserves the pretrained 2D backbone, whether CNN or ViT, and enables 3D fusion by distributing spatial aggregation across orthogonal planes throughout network depth. Specifically, feature interactions are cyclically performed over the $HW$, $DW$, and $DH$ planes across layers, enabling progressive 3D integration in a training-free manner.

We use DINOv3~\cite{simeoni2025dinov3} as a representative. As illustrated in Fig.~\ref{fig:feature_visualization}, slice-wise 2D inference produces features that are well-structured within the selected plane ($HW$) but inconsistent across slices. In contrast, naïvely converted 3D models exhibit weak and poorly aligned 3D representations across all three orthogonal planes prior to retraining. \method{}, however, yields well-aligned 3D features across $HW$, $DW$, and $DH$ without additional supervision. This intrinsic 3D capability is further reflected in linear probing and zero-training evaluations (Sec.~\ref{sec:insights}), where \method{} lifting surpasses 2D/3D by considerable margins. After full fine-tuning, performance remains competitive with full 3D models.

Importantly, \method{} is not a replacement for 3D pretraining~\cite{wang2024internvideo2,hamamci2026generalist,wu2025large} or adapters~\cite{yang2021asymmetric,wu2025medical}, but a complementary operator; lifted models remain fully compatible with these techniques. We aim to demonstrate that 3D capability can be unlocked directly from pretrained 2D foundation models through a simple, seamless, and practical lifting mechanism.

\section{Method}

\begin{figure}[tb]
\centering
\includegraphics[width=.96\textwidth]{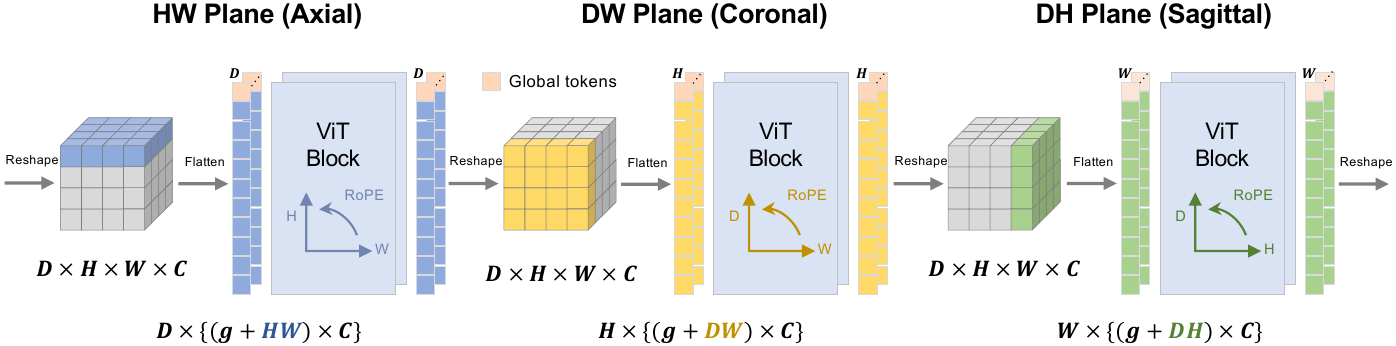}
\caption{Overview of \method{} across three orthogonal planes: $HW$, $DW$, $DH$. Flattened slice tokens are processed by a shared ViT layers with plane-specific RoPE~\cite{su2024roformer}.} \label{fig:Planecycle}
\end{figure}

\subsection{Problem Formulation}
\label{subsec:problem_formulation}

2D-to-3D lifting typically exhibits two extremes: slice-wise 2D processing~\cite{li2025meddinov3,ma2024segment} and full-volume 3D modeling~\cite{wald2503primus,wei2025videorope}. The former extracts features independently from individual slices, whereas the latter operates directly on the full volumes.

For CNNs, 2D-to-3D lifting~\cite{carreira2017quo,yang2021reinventing} primarily increases kernel dimensionality and parameters, while maintaining locally linear complexity. In contrast, Transformer self-attention scales quadratically with sequence length, leading to a substantial increase in attention cost.

Formally, given a 3D feature map $\bXi{} \in \bbR{}^{D\times H\times W\times C}$, the 2D baseline applies pretrained layer $\cFt{}$ independently to each of the $D$ slices. The resulting self-attention cost is $\mathcal{O}((HW)^2)$ per slice, 
leading to a total complexity of $\mathcal{O}(D(HW)^2)$. The 3D baseline flattens the volume and applies $\mathcal{F}_\theta$ once, yielding $\mathcal{O}((DHW)^2)$ self-attention, $D$× higher than 2D slicing.

The 2D baseline preserves computational efficiency but restricts attention to intra-slice tokens. 
The 3D baseline enables global volumetric interaction, yet incurs quadratic self-attention cost. These two baselines reveal a trade-off between efficient 2D weight reuse and scalable volumetric interaction, motivating a unified and computationally efficient lifting strategy.

\subsection{\method{}: Training-Free 2D-to-3D Lifting Operator}
We propose \method{}, an adapter-free operator for 2D-to-3D lifting. 
As detailed in Algorithm~\ref{algo:PC}, the procedure involves plane-wise reshaping, intra-plane aggregation, and 3D restoration. Given a 3D feature map $\bXi$ 
and global tokens $\bGi \in \bbR{}^{P' \times g\times C}$, 
where $g$ denotes the number of global tokens per slice ($g=5$ following DINOv3, consisting of one CLS token and four register tokens), 
we reuse the pretrained 2D weights and select an axis 
$P \in \{D, H, W\}$, corresponding to the $HW$, $DW$, $DH$ plane. 
The volumetric feature map is reshaped into $P$ slices, 
and each slice is flattened into a token sequence of length $DHW/P$, resulting in $\bm{X} = \text{reshape}(\bXi{}) \in \bbR{}^{P \times (DHW/P) \times C}$.
To map global tokens across planes, we apply adaptive average pooling to obtain $\bm{G} = \text{adaptivePool}(\bGi{}) \in \bbR{}^{P \times g \times C} $. The global tokens and patch tokens are concatenated and passed to the frozen 2D layer $P$ times independently, introducing no additional parameters:
\begin{equation}
\bm{T} = \cFt{}\big([\bm{G}, \bm{X}]\big) 
\in \bbR{}^{P \times (g + DHW/P) \times C}.
\end{equation}
After feature aggregation on plane $P$, the global tokens are separated and the patch tokens are reshaped back to the volumetric layout. 
Our method modifies only the viewing plane of the 3D feature representation, introducing negligible computational overhead.

Although the formulation includes global tokens specific to Transformer architectures, the operator is not restricted to Transformers. For CNN backbones, the same procedure applies without global tokens.

\begin{algorithm}[tb]
\caption{\method{}}
\label{algo:PC}
\SetKwInOut{Input}{Input}\SetKwInOut{Output}{Output}\SetKwInOut{Parameter}{Parameter}

\Input{3D feature $\bXi{} \in \bbR{}^{D\times H\times W\times C}$, global tokens $\bGi \in \bbR{}^{P' \times g\times C}$.}

\KwParameter{Pretrained 2D layer $\cFt{}$, plane $P\in\{D, H, W\}$.}

\Output{3D feature $\bXo{} \in \bbR{}^{D\times H \times W \times C}$, global tokens $\bGo{} \in \bbR{}^{P \times g\times C}$.}

\BlankLine

\nl $\bm{X} = \text{reshape}(\bXi{}) \in \bbR{}^{P \times (DHW/P) \times C}$, 
$\bm{G} = \text{adaptivePool}(\bGi{}) \in \bbR{}^{P \times g \times C} $ \;

\nl $\bm{T} = \cFt{} ( [\bm{G},\bm{X}] ) \in \bbR{}^{P \times (g + DHW/P) \times C}$; \tcp{\algcomment{apply 2D layer $P$ times}}

\nl $\bXo{} = \text{reshape}(\bm{T}[ : , g:]) \in \bbR{}^{D \times H \times W \times C}$,
$\bGo{} = \bm{T}[:, :g] \in \bbR{}^{P \times g \times C}$ .

\end{algorithm}

\subsection{Network-Level Lifting}
The \method{} operator preserves input--output structural consistency and can be seamlessly inserted into arbitrary layers of both convolutional and transformer-based architectures without modifying pretrained parameters.
\parag{Cyclic cross-plane feature interaction.}

In principle, the aggregation plane can be scheduled arbitrarily across network depth. 
In practice, we adopt a four-operator cycle:
$HW$ (axial) $\rightarrow$ $DW$ (coronal) $\rightarrow$ $DH$ (sagittal) $\rightarrow$ $HW$.
This schedule follows standard ViT designs defined on $HW$ plane and matches common depths that are multiples of four. Repeating $HW$ within each cycle assigns more capacity to the axial plane, which typically exhibits higher resolution and stronger anatomical continuity in 3D medical data.

Figure~\ref{fig:Planecycle} illustrates the plane-wise operator, showing how feature aggregation transitions from the $HW$ plane to the subsequent $DW$ and $DH$ planes within a single \method{}. The implementation follows Algorithm~\ref{algo:PC}; the last $HW$ operator is omitted for brevity. As DINOv3 incorporates RoPE, each plane-specific operator computes flattened token coordinates for its corresponding plane. This re-indexing introduces no learnable parameters and is inherently compatible with the cyclic plane operators.

\parag{Handling global tokens.}
Since the axis lengths may differ, switching across orthogonal planes induces a token-length mismatch. We adopt a global mean ($PCm$) strategy which averages global tokens across slices and replicates them for the next plane. However, assigning identical tokens to all slices may alter layer-wise token statistics, especially with a frozen backbone. To preserve distribution consistency, we further introduce a grouping variant ($PCg$), averaging tokens within groups for downsampling and broadcasting them for upsampling. Both $PCm$ and $PCg$ can be unified as adaptive average pooling and efficiently implemented using \texttt{AdaptiveAvgPool1d}, which supports arbitrary input and output sizes.

\parag{Complexity analysis.}
We compare the computational complexity of slice-wise 2D, full-volume 3D, and \method{}.
Section~\ref{subsec:problem_formulation} derives the attention cost of the 2D and 3D baselines. For \method{}, the token sequence length depends on the active plane. Assuming cubic volumes with $D = H = W$, each layer processes plane-wise 2D tokens, so the per-layer self-attention complexity matches the slice-wise 2D case, yielding a $D$-fold reduction compared to full 3D attention.

\section{Experiments}

\begin{table}[tb]

\centering
\caption{Linear probing results (averaged over 5 runs) and average training cost on six 3D classification datasets~\cite{medmnistv2}. R denotes fully fine-tuned ResNet-50~\cite{he2016deep}. Best in \textbf{bold}, and those achieved by ours in \colorbox{red!25}{red}.}
\label{tab:cls_LP}

{\fontsize{8}{8}\selectfont
{
\begin{tabular}{lcccccccccccccc}
\toprule
\multirow{2}{*}{Methods}
& \multicolumn{2}{c}{Organ}
& \multicolumn{2}{c}{Nodule}
& \multicolumn{2}{c}{Fracture}
& \multicolumn{2}{c}{Adrenal}
& \multicolumn{2}{c}{Vessel}
& \multicolumn{2}{c}{Synapse} 
& \multicolumn{2}{c}{AVG} \\ 
\cmidrule(lr){2-3}\cmidrule(lr){4-5}\cmidrule(lr){6-7}\cmidrule(lr){8-9}\cmidrule(lr){10-11}\cmidrule(lr){12-13}\cmidrule(lr){14-15}
& AUC & ACC & AUC & ACC & AUC & ACC & AUC & ACC & AUC & ACC & AUC & ACC & AUC & ACC\\
\midrule

R-3D~\cite{medmnistv2} & 99.4 & 88.3 & 87.5 & 84.7 & 72.5 & 49.4 & 82.8 & 74.5 & 90.7 & 91.8 & 85.1 & 79.5 & 86.3 & 78.0 \\ 

R-ACS~\cite{medmnistv2} & 99.4 & 88.9 & 88.6 & 84.1 & 75.0 & 51.7 & 82.8 & 75.8 & 91.2 & 85.8 & 71.9 & 70.9 & 84.8 & 76.2 \\

\midrule

\multicolumn{15}{l}{\textit{ViT-S/16} (PCg Average Training, GPU memory: 2.5 GB; Time: 10 min)} \\

2D~\cite{li2025meddinov3,ma2024segment} & 
98.6 & 84.5 & 89.2 & \textbf{88.4}& 64.1 & 48.2 & \textbf{87.0} & \textbf{83.9} & 84.7 & 88.7 & 85.7 & 83.8& 84.9 & 79.6 \\

3D~\cite{wald2503primus,wei2025videorope} & 
97.4 & 75.0 & 77.7 & 81.5 & \textbf{70.5}& 50.8 & 77.7 & 77.2 & 73.9 & 88.7 & 74.5 & 75.6 & 78.6 & 74.8\\

PCm & 
\cellcolor{red!25}\textbf{99.4}& \cellcolor{red!25}\textbf{90.2}& 88.9 & 86.4 & 62.9 & 45.8 & 85.0 & 83.4 & \cellcolor{red!25}\textbf{87.6}& 88.2 & 85.3 & 83.0 & 84.8 & 79.5 \\

PCg & 
99.2 & 88.2 & \cellcolor{red!25}\textbf{90.7}& 87.0 & 69.6 & \cellcolor{red!25}\textbf{52.0}& 85.9 & 83.3 & 86.2 & \cellcolor{red!25}\textbf{88.8}& \cellcolor{red!25}\textbf{88.6}& \cellcolor{red!25}\textbf{85.3}& \cellcolor{red!25}\textbf{86.7}& \cellcolor{red!25}\textbf{80.8}\\

\midrule

\multicolumn{15}{l}{\textit{ViT-B/16} (PCg Average Training, GPU memory: 4.3 GB; Time: 20 min)} \\
2D~\cite{li2025meddinov3,ma2024segment}& 98.9 & 88.3 & 88.9 & 85.8 & 63.6 & 46.7 & 85.0 & 82.2 & 88.9 & 89.5 & 86.7 & 85.2 & 85.1 & 79.0 \\
3D~\cite{wald2503primus,wei2025videorope} & 97.8 & 77.6 & 81.7 & 83.1 & \textbf{70.5} & \textbf{54.6} & 80.5 & 79.0 & 77.5 & 88.7 & 82.3 & 80.5 & 81.7 & 77.2\\
PCm & \cellcolor{red!25}\textbf{99.8}& \cellcolor{red!25}\textbf{94.2}& 91.0 & 87.4 & 65.5 & 47.1 & 85.6 & 82.6 & 90.3 & 90.6 & 87.4 & 84.9 & 86.6 & 81.2\\
PCg & 99.7 & 93.4 & \cellcolor{red!25}\textbf{92.7}& \cellcolor{red!25}\textbf{88.6} & 69.4 & 51.4 & \cellcolor{red!25}\textbf{86.9} & \cellcolor{red!25}\textbf{83.4} & \cellcolor{red!25}\textbf{90.5} & \cellcolor{red!25}\textbf{90.9} & \cellcolor{red!25}\textbf{88.9} & \cellcolor{red!25}\textbf{85.7} & \cellcolor{red!25}\textbf{87.8} & \cellcolor{red!25}\textbf{82.0}\\
\midrule
\multicolumn{15}{l}{\textit{ViT-L/16} (PCg Average Training, GPU memory: 7.0 GB; Time: 60 min)} \\
2D~\cite{li2025meddinov3,ma2024segment}& 98.2 & 81.5 & 90.1 & \textbf{88.1}& 63.3 & 44.3 & \textbf{86.2} & 82.4 & 81.5 & 88.8 & 86.1 & 81.6 & 84.2 & 77.8
\\
3D~\cite{wald2503primus,wei2025videorope} & 98.6 & 83.3 & 87.5 & 85.6 & \textbf{69.9} & 47.0 & 77.6 & 77.9 & 82.4 & 88.9 & 79.5 & 79.1 & 82.6 & 77.0\\
PCm & \cellcolor{red!25}\textbf{99.5}& \cellcolor{red!25}\textbf{91.5}& \cellcolor{red!25}\textbf{90.9}& 86.8 & 65.6 & 47.4 & 84.0 & \cellcolor{red!25}\textbf{83.3}& 85.2 & 89.2 & 86.7 & 83.4 & 85.3 & 80.2\\
PCg  & 99.2 & 87.7 & \cellcolor{red!25}\textbf{90.9}& 85.5 & 69.3 & \cellcolor{red!25}\textbf{51.7} & 85.2 & 82.6 & \cellcolor{red!25}\textbf{90.1} & \cellcolor{red!25}\textbf{90.0} & \cellcolor{red!25}\textbf{88.9} & \cellcolor{red!25}\textbf{86.5} & \cellcolor{red!25}\textbf{87.3} & \cellcolor{red!25}\textbf{80.7}\\

\bottomrule
\end{tabular}
}
}

\end{table}

\begin{table}[tb]

\centering
\caption{Full fine-tuning (averaged over 5 runs) and average training cost on six 3D classification datasets~\cite{medmnistv2}. Best in \textbf{bold}, and those achieved by ours in \colorbox{red!25}{red}.}
\label{tab:cls_FT}

{\fontsize{8}{8}\selectfont
{
\begin{tabular}{lcccccccccccccc}
\toprule
\multirow{2}{*}{Methods}
& \multicolumn{2}{c}{Organ}
& \multicolumn{2}{c}{Nodule}
& \multicolumn{2}{c}{Fracture}
& \multicolumn{2}{c}{Adrenal}
& \multicolumn{2}{c}{Vessel}
& \multicolumn{2}{c}{Synapse} 
& \multicolumn{2}{c}{AVG} \\ 
\cmidrule(lr){2-3}\cmidrule(lr){4-5}\cmidrule(lr){6-7}\cmidrule(lr){8-9}\cmidrule(lr){10-11}\cmidrule(lr){12-13}\cmidrule(lr){14-15}
& AUC & ACC & AUC & ACC & AUC & ACC & AUC & ACC & AUC & ACC & AUC & ACC & AUC & ACC\\
\midrule

ViViT~\cite{vivit} & 99.9 & 96.5 & 88.3 & 87.2 & 69.9 & 51.0 & 86.3 & 82.9 & 95.1 & 94.0 & 93.5 & 90.5 & 88.8 & 83.7\\

\midrule

\multicolumn{15}{l}{\textit{ViT-S/16} (PCg Average Training, GPU memory: 17 GB; Time: 15 min)} \\
2D~\cite{li2025meddinov3,ma2024segment}& 98.6 & 85.8 & 89.1 & 85.5 & 62.2 & 44.1 & 81.7 & 80.8 & 84.2 & 90.1 & 93.0 & 88.2 & 84.8 & 79.1 \\
3D~\cite{wald2503primus,wei2025videorope} & 99.8 & 95.7 & \textbf{93.8} & \textbf{87.8} & \textbf{76.0} & 55.0 & \textbf{89.7} & \textbf{85.6} & \textbf{94.9} & 91.9 & \textbf{96.5} & \textbf{92.7} & \textbf{91.8} & 84.8 \\
PCm & \cellcolor{red!25}\textbf{99.9} & \cellcolor{red!25}\textbf{96.2} & 93.4 & 87.4 & 73.8 & \cellcolor{red!25}\textbf{56.4} & 89.2 & 85.4 & \cellcolor{red!25}\textbf{94.9} & 93.4 & 95.1 & 91.8 & 91.0 & \cellcolor{red!25}\textbf{85.1}\\
PCg & \cellcolor{red!25}\textbf{99.9} & \cellcolor{red!25}\textbf{96.2} & 93.2 & \cellcolor{red!25}\textbf{87.8} & 72.2 & 51.9 & 88.5 & 84.6 & 94.1 & \cellcolor{red!25}\textbf{93.6} & 95.7 & 92.2 & 90.6 & 84.4 \\
\midrule

\multicolumn{15}{l}{\textit{ViT-B/16} (PCg Average Training, GPU memory: 33 GB; Time: 40 min)} \\
2D~\cite{li2025meddinov3,ma2024segment}& 99.0 & 87.8 & 90.8 & 86.8 & 61.1 & 44.9 & 80.6 & 80.5 & 82.0 & 89.4 & 94.1 & 88.5 & 84.6 & 79.7 \\
3D~\cite{wald2503primus,wei2025videorope} & 99.8 & 96.2 & 92.7 & 87.7 & \textbf{74.2} & 52.6 & 89.0 & \textbf{85.1} & \textbf{94.9} & 92.0 & 96.1 & 91.7 & 91.1 & 84.2 \\
PCm & \cellcolor{red!25}\textbf{100.0} & \cellcolor{red!25}\textbf{97.5} & 91.8 & 88.1 & 72.8 & 52.7 & 86.6 & 81.5 & 94.6 & 91.9 & 96.0 & 92.7 & 90.3 & 84.1 \\
PCg  & 99.9 & 97.0 & \cellcolor{red!25}\textbf{94.6} & \cellcolor{red!25}\textbf{88.5} & 73.1 & \cellcolor{red!25}\textbf{53.1} & \cellcolor{red!25}\textbf{89.1} & 84.8 & 93.8 & \cellcolor{red!25}\textbf{92.7} & \cellcolor{red!25}\textbf{96.9} & \cellcolor{red!25}\textbf{93.3} & \cellcolor{red!25}\textbf{91.2} & \cellcolor{red!25}\textbf{84.9} \\
\midrule

\multicolumn{15}{l}{\textit{ViT-L/16} (PCg Average Training, GPU memory: 87 GB; Time: 120 min)} \\
2D~\cite{li2025meddinov3,ma2024segment} & 98.6 & 84.6 & 90.2 & 86.8 & 64.6 & 43.8 & 81.1 & 80.1 & 85.3 & 89.7 & 90.5 & 86.8 & 85.0 & 78.6 \\
3D~\cite{wald2503primus,wei2025videorope} & 99.8 & 97.2 & 93.6 & \textbf{88.3} & \textbf{76.6} & \textbf{58.4} & \textbf{88.9} & \textbf{84.8} & \textbf{96.7} & \textbf{94.4} & \textbf{97.3} & \textbf{93.0} & \textbf{92.2} & \textbf{86.0} \\
PCm & \cellcolor{red!25}\textbf{99.9} & 97.0 & 93.2 & 86.9 & 73.9 & 57.1 & 87.5 & 83.0 & 95.1 & 93.6 & 96.6 & 91.9 & 91.0 & 84.9 \\
PCg & \cellcolor{red!25}\textbf{99.9} & \cellcolor{red!25}\textbf{97.5} & \cellcolor{red!25}\textbf{94.0} & 87.0 & 73.8 & 53.7 & 87.7 & 83.8 & 95.8 & 92.6 & 97.1 & 90.1 & 91.4 & 84.1 \\

\bottomrule
\end{tabular}
}
}

\end{table}
\begin{figure}[tb]
\centering
\includegraphics[width=.9\textwidth]{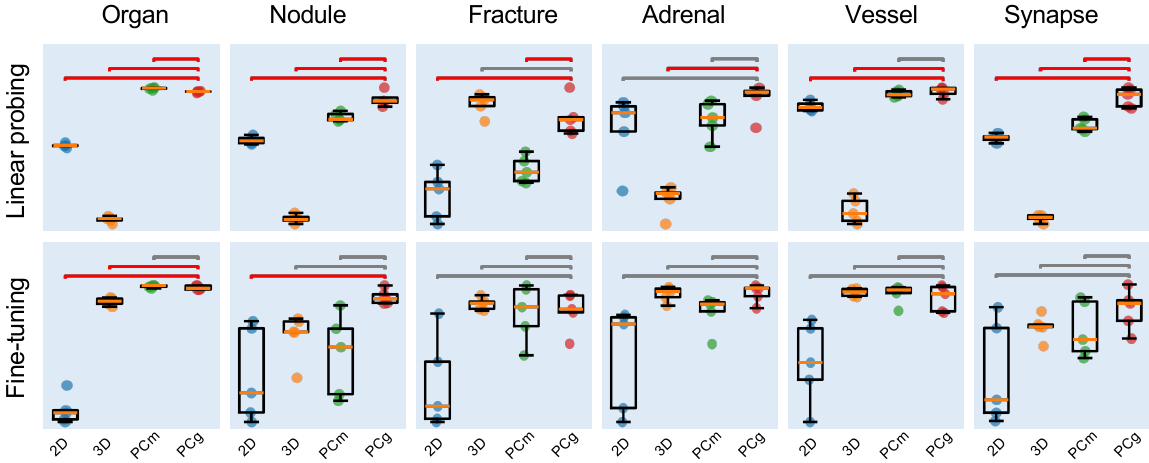}
\caption{Paired t-tests of AUC on six 3D classification datasets~\cite{medmnistv2} on ViT-B/16, computed over five runs. Red indicates significance ($p<0.05$). 
} 
\label{fig:ttest}
\end{figure}


	


\begin{table}[tb]
	\centering
    \caption{Segmentation performance on LIDC~\cite{nodulemnist3d} and MMWHS~\cite{mmwhs} under different settings. Best in \textbf{bold}, and those achieved by ours in \colorbox{red!25}{red}.}

    \label{tab:segmentation-performance}
	\centering
    {
    \fontsize{8}{8}\selectfont
    {
	\begin{tabular}{lcccccccccccc}
		\toprule
         \multirow{3}{*}{Methods} 
         & \multicolumn{4}{c}{ViT-S/16} 
         & \multicolumn{4}{c}{ViT-B/16} 
         & \multicolumn{4}{c}{ViT-L/16} 
         \\
         \cmidrule(lr){2-5}
        \cmidrule(lr){6-9}
        \cmidrule(lr){10-13} 
        & \multirow{2}{*}{LIDC} & \multicolumn{2}{c}{MMWHS}
        & \multirow{2}{*}{AVG}
        & \multirow{2}{*}{LIDC} & \multicolumn{2}{c}{MMWHS} 
        & \multirow{2}{*}{AVG}
        & \multirow{2}{*}{LIDC} & \multicolumn{2}{c}{MMWHS} 
        & \multirow{2}{*}{AVG}
        \\
        &  & CT & MRI &  &  & CT & MRI &  &  & CT & MRI \\
		\midrule

        \multicolumn{13}{l}{\textit{Zero-training Features (FeatDice)}} \\
        \midrule
        2D~\cite{li2025meddinov3,ma2024segment} 
        & 36.9 & \textbf{31.3} & 26.5 & 31.6
        & 28.3 & 30.4 & 26.1 & 28.3
        & 22.4 & 29.1 & 25.1 & 25.5 \\
        
        3D~\cite{wald2503primus,wei2025videorope}
        & 13.9 & 22.4 & 24.4 & 20.2
        & 22.3 & 26.2 & 28.6 & 25.7
        & 21.5 & 26.2 & 27.4 & 25.0  \\
        PCm 
        & 42.3 & \cellcolor{red!25}\textbf{31.3} & \cellcolor{red!25}\textbf{30.8} & \cellcolor{red!25}\textbf{34.8}
        & 30.8 & \cellcolor{red!25}\textbf{33.5} & \cellcolor{red!25}\textbf{29.5} & 31.3
        & 27.9 & \cellcolor{red!25}\textbf{33.7} & \cellcolor{red!25}\textbf{29.2} & \cellcolor{red!25}\textbf{30.3} \\
        PCg 
        & \cellcolor{red!25}\textbf{42.6} & 30.9 & 29.9 & 34.5
        & \cellcolor{red!25}\textbf{32.2} & 33.2 & 29.3 & \cellcolor{red!25}\textbf{31.6}
        & \cellcolor{red!25}\textbf{28.0} & 33.5 & 29.0 & 30.2 \\
        \midrule

        \multicolumn{13}{l}{\textit{Linear Probing}} \\
        \midrule
        2D~\cite{li2025meddinov3,ma2024segment}    	
        & 60.1 & 67.7 & 58.4 & 62.1
        & 62.2 & 68.2 & 57.6 & 62.7
        & 62.6 & 68.8 & 59.5 & 63.6\\
        3D~\cite{wald2503primus,wei2025videorope}
        & 59.4 & 65.9 & 60.9 & 62.1
        & 62.6 & 66.8 & 59.2 & 62.9
        & 61.5 & 65.6 & 57.5 & 61.5\\
        PCm 
        & \cellcolor{red!25}\textbf{64.9} & \cellcolor{red!25}\textbf{69.3} & \cellcolor{red!25}\textbf{61.5} & \cellcolor{red!25}\textbf{65.2}
        & \cellcolor{red!25}\textbf{65.8} & 69.4 & \cellcolor{red!25}\textbf{61.4} & \cellcolor{red!25}\textbf{65.5}
        & \cellcolor{red!25}\textbf{66.1} & \cellcolor{red!25}\textbf{69.2} & 59.8 & 65.0\\
        PCg 
        & 64.4 & 68.7 & 61.4 & 64.8
        & 64.8 & \cellcolor{red!25}\textbf{69.5} & 61.3 & 65.2
        & 65.7 & \cellcolor{red!25}\textbf{69.2} & \cellcolor{red!25}\textbf{61.8} & \cellcolor{red!25}\textbf{65.6}\\
        \midrule

        \multicolumn{13}{l}{\textit{Full Fine-tuning}} \\
        \midrule
        2D~\cite{li2025meddinov3,ma2024segment}
        & 71.2 & 68.4 & 60.7 & 66.8
        & 71.3 & 69.8 & 61.2 & 67.4
        & 71.1 & 70.2 & 61.2 & 67.5\\
        3D~\cite{wald2503primus,wei2025videorope}
        & 72.1 & 71.5 & 66.2 & 69.9
        & 73.4 & 71.0 & \textbf{66.2} & 70.2
        & 74.0 & 72.6 & \textbf{65.5} & 70.7\\
        PCm 
        & 72.6 & \cellcolor{red!25}\textbf{74.2} & 67.8 & 71.5
        & \cellcolor{red!25}\textbf{73.5} & 73.8 & 64.9 & 70.7
        & 73.8 & 74.3 & 64.0 & 70.7\\
        PCg 
        & \cellcolor{red!25}\textbf{72.8} & 74.0 & \cellcolor{red!25}\textbf{70.8} & \cellcolor{red!25}\textbf{72.5}
        & \cellcolor{red!25}\textbf{73.5} & \cellcolor{red!25}\textbf{76.5} & 64.6 & \cellcolor{red!25}\textbf{71.5}
        & \cellcolor{red!25}\textbf{74.4} & \cellcolor{red!25}\textbf{76.2} & 63.8 & \cellcolor{red!25}\textbf{71.5}\\
		\bottomrule
	\end{tabular}
    }
    }

\end{table}

\subsection{Setup}
\paragraph{Datasets.}
We conduct experiments on six 3D datasets: Organ~\cite{organmnist1,organmnist2}, Nodule~\cite{nodulemnist3d}, Fracture~\cite{fracturemnist3d}, Adrenal~\cite{medmnistv2}, Vessel~\cite{vesselmnist3d}, and Synapse~\cite{medmnistv2}. They cover diverse modalities (CT, MRI, and electron microscopy) and anatomical regions (whole-body, lung, abdomen, chest, brain, and cellular structures). We follow the official MedMNIST+~\cite{medmnistv2} preprocessing and split protocol.

For segmentation, we evaluate on LIDC~\cite{nodulemnist3d} for lung nodule segmentation from CT (2,142 for training and 526 for testing) and MMWHS~\cite{mmwhs} for whole heart segmentation from CT and MRI (both use 16 for training and 4 for testing).
\parag{Implementation.}
All experiments are conducted on  a single NVIDIA H200 GPU (141GB). 
We use official pretrained ViT-S/16, ViT-B/16, and ViT-L/16 backbones~\cite{simeoni2025dinov3}. In linear probing, the backbone is frozen and only a classification head or a 3D segmentation decoder is trained. In full fine-tuning, all parameters are optimized.

For classification, models are trained for 10k iterations using AdamW~\cite{loshchilovdecoupled} with 500 warm-up iterations and cosine annealing, and results are averaged over five runs. 
For segmentation, models are trained for 20k iterations using AdamW with 1k warm-up iterations and cosine annealing.
\parag{Metrics.}
We report AUC and ACC for classification, and Dice for segmentation. 

To assess volumetric feature coherence without additional optimization, we define {\em FeatDice}, as no established metric directly evaluates this. For each volume, the feature at the central lesion voxel serves as a reference to compute similarities to all voxel-wise features, yielding a dense similarity map. The Dice coefficient is calculated between the map and the ground-truth lesion mask.
\subsection{Key Insights}
\label{sec:insights}

\paragraph{Proper lifting unlocks the 3D capability of pretrained 2D foundation models.} The proposed adapter-free
\method{} can lift pretrained 2D foundation models for volumetric tasks, even outperforming fully fine-tuned baselines under linear probing. As shown in Tab.~\ref{tab:cls_LP}, $PCg$ surpasses \textit{R-ACS}~\cite{medmnistv2} 3.0 AUC scores and nearly 6.0 ACC scores in average across six benchmarks~\cite{medmnistv2} using ViT-B/16. After fine-tuning (Tab.~\ref{tab:cls_FT}), it further exceeds Transformer-based volumetric models (\textit{ViViT}~\cite{vivit}) by up to 2.6 AUC scores. It indicates that \method{} is more effective preservation of pretrained semantics, whereas \textit{ViViT} lacks comparable pretraining. In same settings, slice-wise 2D processing does not surpass \textit{ViViT}, as it fails to effectively transfer the pretrained model into coherent 3D modeling.


These findings indicate that our method unlocks the 3D capability of pretrained 2D foundation models, even when they are pretrained on natural images.

\parag{Without training, \method{} yields discriminative 3D representations.}
We analyze zero-training features produced from the frozen backbone in Fig.~\ref{fig:feature_visualization} and Tab.~\ref{tab:segmentation-performance}, \method{} produces more coherent feature maps and achieves higher \textit{FeatDice} scores than both 2D and 3D, indicating stronger 3D representations. 

Under linear probing (Tab.~\ref{tab:cls_LP} and Tab.~\ref{tab:segmentation-performance}), \method{} achieves the best performance on nearly all settings. Paired t-tests (Fig.~\ref{fig:ttest}) over five runs further confirm statistical significance ($p<0.05$) on 5/6 datasets under linear probing.


\parag{With fine-tuning, \method{} matches full 3D without compromise.}


Under full fine-tuning (Tab.~\ref{tab:cls_FT} and Tab.~\ref{tab:segmentation-performance}), \method{} consistently outperforms slice-wise 2D processing and achieves performance comparable to 3D flattening. In segmentation, \method{} even surpasses 3D flattening by up to 2.6 Dice points. 
However, 3D flattening requires substantially higher computational cost, incurring a $D\times$ larger self-attention complexity and $>2\times$ the training time (36.2 h vs.\ 16.3 h on ViT-L/16 on LIDC), and performs poorly under linear probing. 

These results indicate that 3D flattening relies heavily on end-to-end fine-tuning to realize its potential, whereas \method{} attains comparable or better performance while retaining 2D's computational efficiency.

\section{Conclusion and Future Work}

We introduced \method{}, a parameter-free operator for architecture-agnostic 2D-to-3D lifting that unlocks 3D capability in pretrained 2D foundation models without architectural modification or retraining, while remaining fully compatible with 3D fine-tuning whenever supervision is available.

Several limitations warrant further investigation. Although \method{} is inherently compatible with both CNN and ViT backbones, we have not yet conducted systematic comparisons. 
The \method{}-lifted models are also applicable to 3D pretraining~\cite{hamamci2026generalist} and adapters such as LoRA~\cite{hu2022lora,liu2025revisiting}, which remain to be explored. 
Besides, while the computational complexity of \method{} matches that of its underlying 2D backbone and suggests favorable scaling behavior compared to full 3D, its large-scale potential has not yet been validated. Preliminary results indicate that lifting DINOv3-7B is feasible, representing a scale not previously achieved in 3D settings.
Moreover, for segmentation, our study intentionally isolates the lifting effect by employing a shared lightweight decoder. The simple decoder and the $16\times$ downsampling in ViT may limit spatial recovery. 

Future work will investigate different architectures, broader datasets, task-specific decoders, and extensions to multimodal or vision–language settings.

\subsubsection{Acknowledgment.}
This study was supported by ELLIS Institute Finland and School of Electrical Engineering, Aalto University. We acknowledge CSC-IT Center for Science, Finland, for providing access to the supercomputers Mahti and Puhti, as well as LUMI, owned by the European High Performance Computing Joint Undertaking (EuroHPC JU) and hosted by CSC Finland in collaboration with the LUMI consortium. We also acknowledge the computational resources provided by the Aalto Science-IT project through the Triton cluster.

\subsubsection{Disclosure of Interests.}
The authors have no competing interests to declare that are relevant to the content of this article.

\bibliographystyle{splncs04}
\bibliography{bib/string,bib/reference}

 \end{document}